\documentclass[10pt,twocolumn,letterpaper]{article}

\usepackage{cvpr}            % Camera-ready
\usepackage{array}

\definecolor{cvprblue}{rgb}{0.21,0.49,0.74}
\usepackage[
    pagebackref,
    breaklinks,
    colorlinks,
    allcolors=cvprblue
]{hyperref}

\usepackage{tabularx} % test

\def\paperID{*****}
\def\confName{CVPR}
\def\confYear{2026}

\title{
Post-Launch Capability Expansion of Vision–Language Models via Prompting for On-Orbit Spacecraft Inspection
}

\author{
\makebox[\textwidth][c]{%
\begin{tabular}[t]{ccc}
\begin{tabular}[t]{c}
Nicholas A. Welsh$^{\dagger}$\\
Florida Institute of Technology\\
Melbourne, FL, USA\\
{\tt\small nwelsh2024@my.fit.edu}
\end{tabular}
&
\begin{tabular}[t]{c}
Lennon Shikhman$^{\dagger}$\\
Florida Institute of Technology\\
Melbourne, FL, USA\\
{\tt\small lshikhman2022@fit.edu}
\end{tabular}
&
\begin{tabular}[t]{c}
M. Nehru Attazs\\
Florida Institute of Technology\\
Melbourne, FL, USA\\
{\tt\small mattzs2012@my.fit.edu}
\end{tabular}
\\[1em]
\begin{tabular}[t]{c}
Seemanthini K. Putane\\
Florida Institute of Technology\\
Melbourne, FL, USA\\
{\tt\small seemanthinipk@gmail.com}
\end{tabular}
&
\begin{tabular}[t]{c}
V. Minh Nguyen\\
University of Florida\\
Gainesville, FL, USA\\
{\tt\small mnguyen2018@my.fit.edu}
\end{tabular}
&
\begin{tabular}[t]{c}
Ryan T. White\\
Florida Institute of Technology\\
Melbourne, FL, USA\\
{\tt\small rwhite@fit.edu}
\end{tabular}
\end{tabular}%
}
}

\begin{document}
\maketitle

%\UNCOMMENT FOR CAMERA READY
\footnotetext{$\dagger$Equal contribution (co-first authors).}

% Abstract
\begin{abstract}
Spaceborne inspection systems often deploy perception models prior to launch, after which updating model weights or expanding fixed label sets becomes operationally impractical. While supervised models can be integrated pre-flight, adding new semantic capabilities in orbit requires retraining and re-uploading parameters. We investigate whether prompt-driven vision–language models can enable post-launch semantic expansion, allowing new spacecraft components to be specified via natural-language prompts without modifying onboard weights. We evaluate zero-shot instance segmentation of spacecraft components under a strictly frozen, single-pass inference protocol on a test set of 129 images of previously unseen satellites. Under fixed global thresholds and no post-processing, SAM3 achieves 0.385 mAP@0.5 and 0.267 mAP@0.5:0.95. Performance is strongly scale-dependent: large structural elements like spacecraft bodies (0.639 AP@0.50) and solar arrays (0.598 AP@0.5) localize reliably, while relatively small appendages like antennas (0.221 AP@0.5) and thrusters (0.081 AP@0.5) remain difficult. Prompt formulation influences performance, with structured prompts incorporating spatial and geometric descriptors yielding up to 82\% improvement over short category-name prompts. The model operates within the memory and compute envelope of contemporary embedded GPUs, suggesting prompt-driven grounding can provide a practical mechanism for post-launch semantic extension of dominant spacecraft structures while highlighting limitations of zero-shot localization for fine-scale components under orbital domain shift.
\end{abstract}

% Main Paper
\section{Introduction}

Autonomous rendezvous and on-orbit inspection rely on visual perception to identify spacecraft structures, assess configuration, and support proximity operations. In operational settings, onboard compute is constrained and flight software stacks are tightly controlled, making post-launch retraining or multi-gigabyte weight updates impractical. We consider a deployment scenario in which a pretrained vision–language model (VLM) is integrated prior to launch and must operate with frozen weights throughout the mission. Once in orbit, adaptation is limited to lightweight inputs such as natural-language prompts.

Promptable VLMs offer a potential mechanism for extending semantic capability under this constraint. Rather than expanding a fixed label set through retraining, these models condition localization directly on textual descriptions. In principle, this allows new inspection objectives to be specified via byte-scale prompt uplink without modifying onboard parameters. The central question is whether such zero-shot grounding remains reliable under the geometry, reflective texture, scale variation, and viewpoint shifts characteristic of orbital inspection imagery.

We therefore evaluate prompt-driven, zero-shot instance segmentation of spacecraft components under a strictly frozen, single-pass inference protocol. By disallowing task-specific fine-tuning, interactive refinement, and post-processing, we isolate the behavior of text-conditioned grounding under deployment-faithful assumptions. The objective is to assess practical reliability rather than propose architectural modifications.

Our results reveal a consistent scale-dependent asymmetry: large structural elements can be localized reliably under fixed prompts, while fine-scale appendages remain substantially more challenging. Prompt formulation materially influences performance, indicating that semantic phrasing plays a central role in zero-shot deployment regimes. These findings clarify both the viability and the limitations of prompt-driven localization as a post-launch adaptation mechanism for spaceborne inspection systems.

\definecolor{clrbody}{RGB}{255,100,200}
\definecolor{clrsolar}{RGB}{0,200,220}
\definecolor{clrantenna}{RGB}{255,165,0}
\definecolor{clrthruster}{RGB}{220,0,0}

\begin{figure*}[t]
\centering

\begin{subfigure}[t]{0.24\textwidth}
\centering
\includegraphics[width=\linewidth]{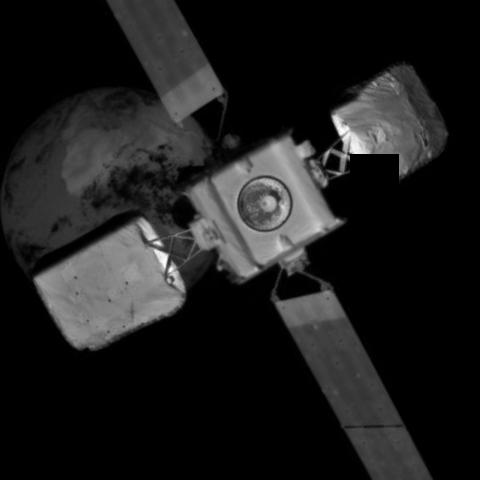}
\caption{Input Image}
\end{subfigure}%
\hfill
\begin{subfigure}[t]{0.24\textwidth}
\centering
\includegraphics[width=\linewidth]{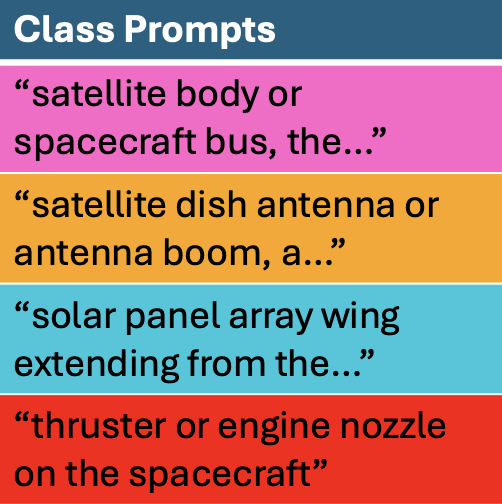}
\caption{Prompts to Frozen SAM3}
\end{subfigure}%
\hfill
\begin{subfigure}[t]{0.24\textwidth}
\centering
\includegraphics[width=\linewidth]{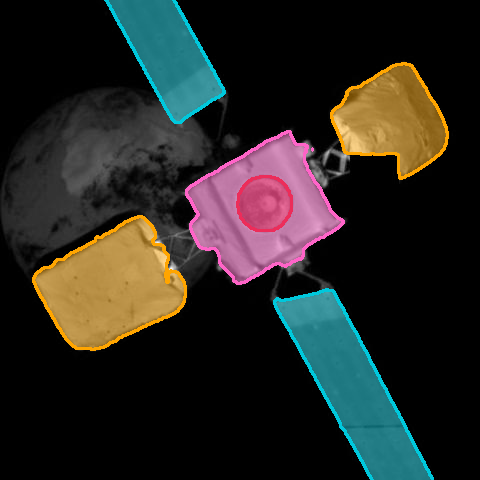}
\caption{Prediction Overlay}
\end{subfigure}%
\hfill
\begin{subfigure}[t]{0.24\textwidth}
\centering
\includegraphics[width=\linewidth]{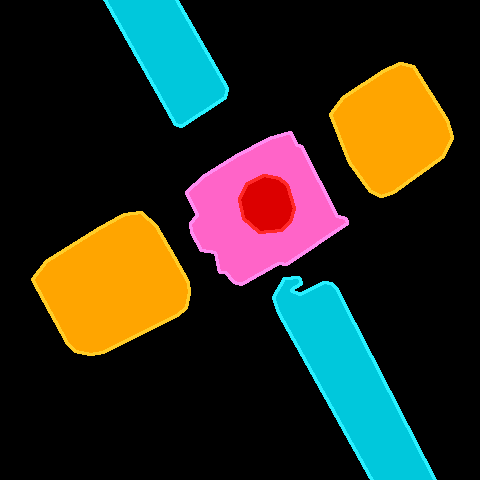}
\caption{Ground-Truth Masks}
\end{subfigure}

\caption{
Prompt-driven spacecraft component segmentation under frozen deployment constraints. A pretrained SAM3 model is assumed to be launched with fixed weights; new inspection objectives are specified post-launch via lightweight natural-language prompts, without retraining or parameter updates. Instance-level masks are produced in a single forward pass. Shown is a held-out test image with predictions for four fixed prompts (Table~\ref{tab:prompts}). Colors: \textcolor{clrbody}{spacecraft body (pink)}, \textcolor{clrsolar}{solar array (cyan)}, \textcolor{clrantenna}{antenna (orange)}, \textcolor{clrthruster}{thruster (red)}.
}
\label{fig:qualitative}
\end{figure*}
\section{Problem Formulation}

We study text-conditioned instance segmentation of spacecraft components in on-orbit inspection imagery. Given an input image and a fixed natural-language prompt describing a component category (e.g., \textit{antenna}, \textit{thruster}, \textit{solar array}, or \textit{spacecraft body}), the objective is to produce one or more segmentation masks corresponding to regions consistent with that description.

Evaluation is conducted in a strictly frozen, zero-shot regime: model parameters are fixed and prompts are defined once per class and applied uniformly across all images. The pretrained VLM receives no spacecraft-specific supervision, and performance is measured on unseen satellite configurations to assess whether prompt conditioning alone can extend semantic coverage without retraining.

\section{Related Work}

\paragraph{Open-vocabulary detection and segmentation.}
Open-vocabulary object detection conditions localization on language embeddings rather than fixed category heads, typically using pretrained VLMs. OWL-ViT demonstrated text-conditioned zero-shot detection via contrastive image–text alignment \cite{minderer2022simpleopenvocabularyobjectdetection}, and subsequent work improved performance through scaling and grounding strategies \cite{minderer2024scalingopenvocabularyobjectdetection, liu2024groundingdinomarryingdino}. Promptable segmentation models such as SAM \cite{kirillov2023segment} and its language-conditioned variants \cite{carion2025sam3segmentconcepts} decouple mask prediction from fixed semantic classes, enabling text-driven segmentation. These methods are primarily evaluated on natural-image benchmarks under unconstrained inference settings.

\paragraph{Spacecraft perception and domain shift.}
Existing spacecraft perception systems are typically closed-set and supervised for mission-specific targets \cite{Yan2018SpacecraftDB, mahendrakar_real-time_2024, Park_2025, 10115705}. In contrast, open-vocabulary grounding methods are trained predominantly on natural-image data and may degrade under domain shift. We evaluate a distinct regime: prompt-driven, zero-shot instance segmentation of spacecraft components under a strictly frozen, single-pass deployment protocol on previously unseen satellites.

\paragraph{Positioning of this work.}
Rather than proposing a new architecture, we assess whether prompt conditioning alone can extend the semantic scope of a generic pretrained model to spacecraft inspection imagery without retraining or weight updates. This isolates the practical viability and limitations of prompt-driven semantic expansion in spaceborne deployment scenarios.

\section{Datasets}

Images are drawn from the Web Satellite Dataset (WSD) \cite{mahendrakar_real-time_2024}, originally annotated for object detection. For this study, we augment WSD with instance-level segmentation masks for four spacecraft component classes: antenna, thruster, solar array, and spacecraft body. The dataset contains 1,433 images (8,141 instances) in a development split and 129 images (871 instances) in a held-out test split.

\section{Methods}

\paragraph{Model.} We evaluate Segment Anything Model 3 (SAM3) \cite{carion2025sam3segmentconcepts}, a promptable vision–language segmentation model. All experiments use SAM3 in a strictly zero-shot setting with fully frozen weights; no fine-tuning, adapter layers, or parameter updates are performed.

\paragraph{Prompt Design.} Each class is assigned a single fixed natural-language prompt applied uniformly across all images. Prompts describe visual appearance and spatial context rather than relying solely on category names. Table~\ref{tab:prompts} lists the final prompts used for all reported results.

\begin{table}[t]
\centering
\caption{Fixed prompts used for each spacecraft component class. Prompts were selected on the development split prior to held-out test evaluation.}
\label{tab:prompts}
\small
\begin{tabularx}{\columnwidth}{lX}
\toprule
Class & Prompt \\
\midrule
Spacecraft Body &
``satellite body or spacecraft bus, the main boxy or cylindrical structure'' \\

Antenna &
``satellite dish antenna or communication antenna boom, a circular parabolic reflector or thin rod extending outward from the spacecraft'' \\

Solar Array &
``solar panel array wing extending from the spacecraft'' \\

Thruster &
``thruster or engine nozzle on the spacecraft'' \\
\bottomrule
\end{tabularx}
\end{table}

Prompt variants were compared on the development split. Three consistent effects emerged. First, compound subtype descriptions improved recall for classes with multiple morphologies (e.g., including ``circular parabolic reflector'' and ``thin rod'' in the antenna prompt). Second, explicit spatial grounding phrases such as ``extending from the spacecraft'' disambiguated visually similar structures. For solar arrays, adding spatial context improved AP@0.5 from 0.367 to 0.580. Third, geometric descriptors such as ``boxy or cylindrical'' improved precision for structurally ambiguous components, yielding up to 82\% AP@0.5 improvement for spacecraft body. Overly verbose prompts reduced recall, and more expensive multi-prompt ensembles degraded mAP by increasing false positives without recall gains.

\section{Experimental Results}

\paragraph{Evaluation Protocol.}
All experiments are conducted in a strictly frozen, zero-shot setting with single-pass inference. Each image–prompt pair is processed once without interactive refinement, and model weights remain fixed throughout. Ground-truth annotations are used solely for offline evaluation. Prompts and inference thresholds are fixed prior to held-out test evaluation.

Instance segmentation performance is measured using per-class average precision (AP) and mean average precision (mAP) across classes. We report each at both 0.5 threshold and averaged over IoU thresholds from 0.5 to 0.95 in increments of 0.05 following standard COCO-style evaluation. They are denoted AP@0.5 and AP@0.5:0.95.

Predicted masks are binarized using a fixed mask threshold of 0.40 and filtered with a fixed score threshold of 0.05. No class-specific tuning or morphological post-processing is applied. All thresholds are fixed uniformly across classes and images.

\paragraph{Instance-level segmentation performance.}

We evaluate zero-shot, prompt-driven segmentation on the held-out test set (129 images, 871 instances), which was not used during prompt development. Table~\ref{tab:instance_results} reports per-class AP and overall mAP. Under fixed global thresholds and without post-processing, SAM3 achieves 0.385 mAP@0.5 and 0.267 mAP@0.5:0.95.

\newcolumntype{Y}{>{\raggedright\arraybackslash}X}
\begin{table}[b]
\centering
\caption{Instance-level zero-shot performance on the held-out test set (129 images, 871 instances).}
\label{tab:instance_results}
\begin{tabularx}{\columnwidth}{@{}Ycc@{}}
\toprule
Class & AP@0.5 & AP@0.5:0.95 \\
\midrule
Antenna         & 0.221 & 0.133 \\
Thruster        & 0.081 & 0.043 \\
Solar Array     & 0.598 & 0.439 \\
Spacecraft Body & 0.639 & 0.453 \\
\midrule
Overall (mAP)   & 0.385 & 0.267 \\
\bottomrule
\end{tabularx}
\end{table}

Performance exhibits a clear scale-dependent asymmetry. Large structural components like spacecraft bodies (0.639 AP@0.5) and solar arrays (0.598 AP@0.5) localize reliably with stable mask quality across IoU thresholds. In contrast, smaller components remain substantially more challenging: antennas (0.221 AP@0.5) and thrusters (0.081 AP@0.5). Performance on the test set closely matches development-split behavior, indicating that prompt improvements transfer without significant degradation.

\paragraph{Effect of prompt formulation.}
Prompt variants were compared on the development split prior to held-out evaluation. Replacing structured prompts with short category-name prompts substantially reduced performance. For example, solar-array AP@0.5 improved from 0.367 to 0.580 (+58\%) when spatial context was added, and spacecraft-body AP@0.5 improved from 0.267 to 0.485 (+82\%) when geometric descriptors were included. Across classes, prompt formulation produced larger performance gains than threshold adjustment or post-processing, indicating that semantic phrasing materially shapes zero-shot grounding behavior in frozen deployment regimes.
\section{Analysis and Discussion}
\paragraph{Grounding performance is strongly scale-dependent.} Localization exhibits a consistent scale-dependent asymmetry: large structural components (spacecraft bodies and solar arrays) are grounded reliably under fixed prompts, while small appendages remain difficult. Large planar or boxy structures provide strong spatial and contextual cues that align well with open-vocabulary grounding. In contrast, thin antenna booms and small propulsion elements occupy limited pixel support and often appear at low contrast. Thrusters are particularly challenging because their appearance varies across viewpoint and scale, limiting the effectiveness of a single fixed prompt.

\paragraph{Prompt wording drives zero-shot grounding behavior.} Across experiments, prompt design produced larger performance changes than threshold adjustment or post-processing. Structured prompts incorporating compound descriptions, spatial context (e.g., ``extending from the spacecraft''), and geometric descriptors (e.g., ``boxy or cylindrical'') substantially improved localization relative to short category-name prompts. In frozen, zero-shot regimes, semantic phrasing alone influences grounding behavior without weight modification.

\paragraph{Failures concentrate on small and visually ambiguous structures.} Two recurring failure patterns were observed. First, at low score thresholds, the model produces many low-confidence candidate masks, leading to over-segmentation. Second, for small components, visually similar structures are frequently mislocalized (e.g., antenna booms confused with structural struts). These errors reflect limited discriminative signal at small spatial scales rather than systematic semantic confusion.

\paragraph{Prompt-driven grounding is viable for dominant structural elements.} Under deployment constraints where retraining is infeasible, prompt-driven grounding appears practical for identifying dominant structural components of a spacecraft. However, reliable detection of fine-scale appendages will likely require additional supervision, domain adaptation, higher-resolution sensing, or complementary modalities. The sensitivity to prompt wording suggests that deployment should include structured prompt validation prior to flight.

\paragraph{Implications for deployed spaceborne AI systems.}
Traditional spacecraft perception pipelines treat semantic scope as fixed at launch: expanding the set of recognizable components typically requires retraining, revalidation, and uploading updated model parameters. In contrast, prompt-driven grounding decouples semantic specification from parameter updates. Under this paradigm, new inspection targets can be introduced post-launch through lightweight textual uplink rather than multi-gigabyte weight replacement. This distinction is operationally significant in spaceflight contexts, where bandwidth is limited, software modifications trigger costly recertification processes, and onboard compute resources are tightly managed. Inference, however, remains compatible with contemporary embedded GPU platforms, making this extension mechanism software-centric rather than hardware-dependent.

Our results suggest that such semantic extension is practically viable for dominant structural components, even under strict frozen-weight constraints and single-pass inference. While performance degrades for fine-scale appendages, the demonstrated reliability for large spacecraft structures indicates that prompt-based adaptation can meaningfully expand the functional envelope of a deployed model without altering its parameters. This reframes prompt engineering as a practical mechanism for extending the functional scope of on-board AI systems after launch.

\section{Conclusion}

\paragraph{Zero-shot performance under frozen deployment.} We evaluated prompt-driven, zero-shot instance segmentation of spacecraft components under a frozen model, single-pass inference protocol. Without retraining or post-processing, SAM3 reliably localized dominant structural components with reliable accuracy on previously unseen satellites. Small components such as antennas and thrusters remained substantially more challenging. Performance exhibited a consistent scale-dependent asymmetry, with large components approaching practical reliability and small components degrading rapidly under stricter localization criteria.

\paragraph{Prompt formulation as semantic adaptation.} Prompt formulation materially influenced performance. Structured prompts incorporating spatial context and geometric descriptors led to increased performance compared to class-name prompts, demonstrating that semantic phrasing meaningfully shapes grounding behavior under frozen-weight constraints. In this setting, prompt design serves as a lightweight mechanism for adapting semantic objectives using byte-scale short text prompts rather than multi-gigabyte weight uplinks.

\paragraph{Scope and limitations.} Evaluation is conducted on a single spacecraft imagery dataset covering four component types, and only one promptable VLM is studied. Findings therefore characterize zero-shot behavior of this model under frozen, single-pass inference and should not be interpreted as fundamental limits of open-vocabulary grounding.. Runtime and energy characteristics on flight-relevant hardware are not directly measured. Additionally, the observed sensitivity to prompt wording indicates that prompt selection itself constitutes a form of task calibration, even in the absence of parameter updates.

\paragraph{Practical deployment implications.} Practically, prompt-driven grounding appears viable for identifying dominant structural components under deployment-motivated constraints and can plausibly operate within the compute envelope of modern embedded GPUs (e.g., Jetson Orin–class devices). However, reliable localization of small components is unlikely to reach operational thresholds with this approach. Achieving operational reliability for small or visually ambiguous components will likely require domain-specific pretraining or supervised adaptation on representative spacecraft imagery prior to deployment.

% References
{
    \small
    \bibliographystyle{plainnat}
    \bibliography{main}
}

% Acknowledgements
\section*{Acknowledgements}

\paragraph{Computational Resources.} The authors gratefully acknowledge Dell Technologies, and in particular the Dell Pro Precision division, for providing computational resources that supported the experiments in this work. All experiments were conducted on a Dell Pro Max T2 workstation equipped with an Intel Core Ultra 9 285K processor, 128 GB of DDR5 ECC memory, and an NVIDIA RTX PRO 6000 Blackwell GPU. The views and conclusions expressed herein are those of the author and do not necessarily reflect the views of Dell Technologies.

% \paragraph{Reproduciblity.} Code to reproduce all experiments, generate figures, and evaluate results is available at \url{https://github.com/lennonshikhman/vlm_sat_project}. The repository includes implementation details, training and evaluation scripts, and instructions necessary to replicate the reported results.

\end{document}